%% file: neurips_data_2024.tex
\documentclass{article}
\usepackage[dvipsnames]{xcolor}

\usepackage[preprint]{neurips_data_2024}

\usepackage[utf8]{inputenc} 
\usepackage[T1]{fontenc}    
\usepackage{hyperref}       
\usepackage{url}            
\usepackage{booktabs}       
\usepackage{amsfonts}       
\usepackage{nicefrac}       
\usepackage{microtype}      
\usepackage{xcolor}         
\usepackage{verbatim}
\usepackage{natbib}
\usepackage{booktabs}
\usepackage{multirow}
\usepackage{array}
\usepackage{graphicx}
\usepackage[utf8]{inputenc} 
\usepackage[T1]{fontenc}    
\usepackage{hyperref}       
\usepackage{url}            
\usepackage{booktabs}       
\usepackage{amsfonts}       
\usepackage{nicefrac}       
\usepackage{microtype}      
\usepackage{xcolor}         
\usepackage{dirtree}
\usepackage{bbding}
\usepackage{multirow}
\usepackage{listings}
\usepackage{graphicx}
\usepackage{framed}
\usepackage[most]{tcolorbox}
\usepackage{fontawesome5}
\usepackage{caption}
\usepackage{lipsum}
\usepackage{adjustbox}
\usepackage{enumitem}
\usepackage{colortbl}

\newcommand{\grayrow}{\rowcolor[gray]{.9}}
\definecolor{baselinecolor}{gray}{.95}

\colorlet{darkgreen}{green!65!black}
\definecolor{redboxcolor}{rgb}{0.6, 0, 0}
\definecolor{greenboxcolor}{rgb}{0.42, 0.66, 0.31}
\definecolor{purpleboxcolor}{rgb}{0.56, 0.49, 0.77}

\title{WeatherQA: Can Multimodal Language Models Reason about Severe Weather?}

\author{
  Chengqian Ma$^{1}$\thanks{Equal Contribution}\;\;
  Zhanxiang Hua$^{2}$\footnotemark[1]\;\;
  Alexandra Anderson-Frey$^2$ \;\; \\
  \textbf{Vikram Iyer}$^3$ \;\;
  \textbf{Xin Liu}$^3$\;\;
  \textbf{Lianhui Qin}$^{4,5}$ \\
    $^1$Department of Applied Mathematics, University of Washington \\
    $^2$Department of Atmospheric Sciences, University of Washington \\
    $^3$Allen School of Computer Science \& Engineering, University of Washington\\
    $^4$University of California, San Diego~~ $^5$Allen Institute for Artificial Intelligence
}

\begin{document}

\maketitle

\begin{abstract}
Severe convective weather events, such as hail, tornadoes, and thunderstorms, often occur quickly yet cause significant damage, costing billions of dollars every year. This highlights the importance of forecasting severe weather threats hours in advance to better prepare meteorologists and residents in at-risk areas. Can modern large foundation models perform such forecasting? Existing weather benchmarks typically focus only on predicting time-series changes in certain weather parameters (e.g., temperature, moisture) with text-only features. In this work, we introduce WeatherQA, the first \textit{multimodal} dataset designed for machines to reason about complex \textit{combinations} of weather parameters (a.k.a., \textit{ingredients}) and predict severe weather in real-world scenarios. The dataset includes over 8,000 (multi-images, text) pairs for diverse severe weather events. Each pair contains rich information crucial for forecasting---the images describe the ingredients capturing environmental instability, surface observations, and radar reflectivity, and the text contains in-depth forecast analyses written by human experts. With WeatherQA, we systematically evaluate state-of-the-art vision language models (VLMs), including GPT4, Claude3.5, Gemini-1.5, and a fine-tuned Llama3-based VLM, by designing two challenging tasks: (1) multi-choice QA for predicting affected area and (2) classification of the development potential of severe convection. 
These tasks require deep understanding of domain knowledge (e.g., atmospheric dynamics) and complex reasoning over multimodal data (e.g., interactions between weather parameters). We show a substantial gap between the strongest VLM, GPT4o, and human reasoning. Our comprehensive case study with meteorologists further reveals the weaknesses of the models, suggesting that better training and data integration are necessary to bridge this gap. WeatherQA is accessible through \url{https://github.com/chengqianma/WeatherQA}.
\end{abstract}

\setcitestyle{numbers}
\section{Introduction}
Severe convective weather events, including hail, tornadoes, damaging winds and thunderstorms, pose significant risks to human life, infrastructure, and property, causing considerable losses every year. The United States alone reports total economic losses exceeding \$10 billion each year, while Europe experiences losses between €1–2 billion. Moreover, these events are increasing in damage \cite{hoeppe2016trends,franzke2021towards, newman2023global}, emphasizing the importance of analyzing severe weather threats in real-time and promptly communicating the information to both forecasters and people living in affected regions.

\begin{figure}[t]
    \vskip -0.13in
    \centering
    \includegraphics[width=\textwidth, height=\textheight, keepaspectratio]{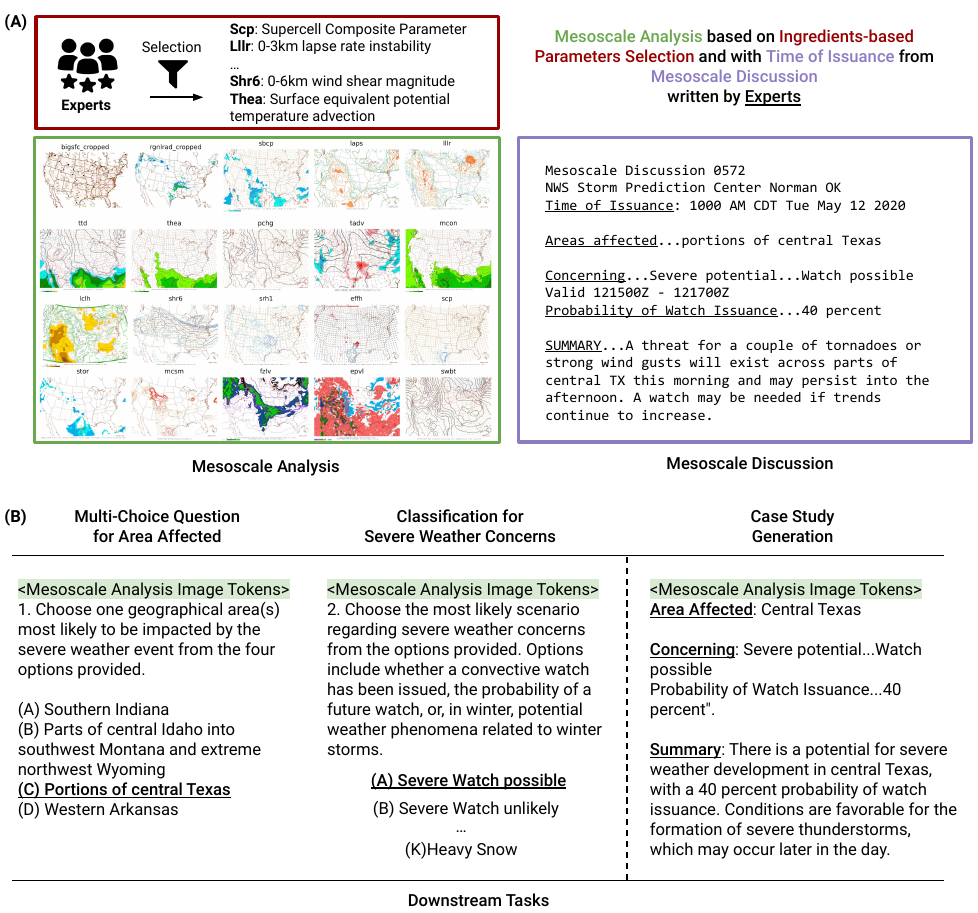}
    \vspace{-15pt}
    \caption{\textbf{An Overview of WeatherQA.} (A): The data curation process involves selecting ingredient-based parameters by forecasters and pairing them with the date and hour from the Mesoscale Discussion to extract the corresponding Mesoscale Analysis. (B): We introduce two downstream tasks which are identifying regions with potential severe weather threats and identifying the types of severe weather concerns issued in a Mesoscale Discussion. A case study investigates the VLM-generated output of Mesoscale Discussions reviewed by experts.}
    \label{fig:weatherqa_concept}
    \vspace{-15pt}
\end{figure}

We introduce WeatherQA, the first multimodal dataset and benchmarks designed to improve and evaluate AI models for reasoning about severe weather in real-world scenarios. 
Collected and processed from the NOAA Storm Prediction Center \footnote{\url{www.spc.noaa.gov/products/md/} and \url{www.spc.noaa.gov/exper/ma_archive/}}, the dataset includes 8,511 (multi-image, text) pairs of weather data from severe convective and winter storms together with expert analyses.

While traditional weather tasks and datasets focus only on predicting time-series changes in individual weather parameters (such as temperature and humidity), expert meteorologists forecast weather events based complex {\it ingredients}, i.e., combinations of key weather parameters such as moisture, lift, instability ({\it Lllr}), and wind shear ({\it Shr6}) \cite{brooks2007ingredients} (Figure \ref{fig:weatherqa_concept}(A), \textcolor{redboxcolor}{red} box). 

More specifically, the expert meteorologists integrate information such as surface observations, radar reflectivity, and ingredients-based parameters (Figure \ref{fig:weatherqa_concept}(A), \textcolor{greenboxcolor}{green} box) to write Mesoscale Discussions (MCDs) that describe the potential of specific severe weather threats hours in advance (Figure \ref{fig:weatherqa_concept}(A), \textcolor{purpleboxcolor}{purple} box). 
Accordingly, each example of the WeatherQA dataset consists of (1) multiple images describing a standardized set of ingredients-based parameters, composite radar reflectivity, and surface observations over the contiguous United States (CONUS), and (2) the corresponding MCD. 

To the best of our knowledge, WeatherQA is the first dataset integrating multimodal weather data with expert-level text reasoning.

As shown in Figure \ref{fig:weatherqa_concept}(B), WeatherQA also designs two challenging tasks to systematically test current AI models. The first task is multiple-choice QA which involves predicting the affected area based on the current time and weather conditions described by ingredients-based weather parameters (Figure \ref{fig:weatherqa_concept}(B), left). The second task is classification which focuses on determining the development potential of severe convection based on domain knowledge about the weather parameters and radar reflectivity (Figure \ref{fig:weatherqa_concept}(B), middle). These tasks are particularly challenging as they require: (1) weather domain knowledge including complex interconnections between atmospheric parameters that vary in space and time, (2) understanding of the complexities of surface geographical features and their interactions with the atmosphere, and (3) the ability to reason toward an outcome given diverse pieces multimodal information, such as interpreting the spatial movement of precipitation patterns based on radar reflectivity and wind field. 

Modern large vision-language models (VLMs) \cite{liu2024llava, zhang2024ferret} have shown promise in integrating visual and textual data. We conduct the first comprehensive analysis of VLMs on severe weather reasoning. Specifically, we evaluate state-of-the-art VLMs on WeatherQA, including GPT4, Gemini1.5, Claude3.5, and a Llama3-tuned VLM, on the above two tasks and compare against the ground truth from the NOAA Storm Prediction Center discussion notes. Empirical results indicate a substantial performance gap between VLMs and human reasoning. Specifically, VLMs struggle to consider the relations between ingredients-based weather parameters and properly refer back to the geographical regions over the CONUS. Furthermore, VLMs lack domain knowledge in evaluating the severity of weather events, even when they correctly identify geographical regions (e.g., GPT-4o overestimates the watch issuance probability as shown in Figure~\ref{fig:case study demo}, \textcolor{purpleboxcolor}{purple} text). 

For deeper insights into the model behaviors on WeatherQA, we conduct a comprehensive case study with human experts (Figure \ref{fig:weatherqa_concept}(B), right). This study analyzes the performance of VLMs in generating the reasoning process of weather predictions. Interestingly, in addition to identifying limitations such as incorrect regional placements relative to potential hazards and erroneous hazard identification as above, the analysis also recognizes instances where the models provided valuable complementary information that expert forecasters had overlooked.

WeatherQA aims to facilitate the development of more accurate and reliable AI-driven weather reasoning models. To support the community, we will publicly release the following assets: the (multi-image, text) pair weather analysis dataset, codebase for data curation, finetuned models, evaluation scripts, as well as the expert-reviewed rubric, guidelines, and feedback on sample answers.

\section{Related Work}

{\bf Deep Learning Numerical Model for Weather and Climate } 
Recent advances in AI for weather and climate are revolutionizing weather forecasting, climate simulation \cite{watt2023ace}, and atmospheric downscaling \cite{li2024corrdiff}.  Models like FourCastNet \citep{pathak2022fourcastnet}, Pangu-Weather \citep{bi2023accurate}, GraphCast \citep{lam2023learning}, FuXi \citep{chen2023fuxi}, FengWu \citep{chen2023fengwu}, and Stormer \citep{nguyen2023scaling}, are challenging state-of-the-art numerical prediction models. Foundation models trained for weather and climate like Aurora \cite{bodnar2024aurora} are advancing rapidly. WeatherQA focuses on the next stage of the weather prediction when model outputs are used to create forecasts and recommendations.  

{\bf Large Language Models for Climate Text } 
Large language models (LLMs) have shown progress in climate text-based-only analysis \cite{chen2023foundation, zhu2024foundations}. Climate-specific models like ClimateBERT \cite{webersinke2021climatebert}, based on DistilROBERTA \cite{sanh2019distilbert}, are used for tasks like detecting climate-related content and sentiment analysis. Extensions include ClimateBERT-NetZero \cite{schimanski2023climatebert} for classifying net zero targets, and ClimateNLP \cite{anoop2023climatenlp} for analyzing public sentiment on social media. Researchers have explored pre-trained LLMs' zero-shot capabilities for evaluating TCFD reporting \cite{auzepy2023evaluating}, and integrated emission data to enhance models \cite{auzepy2023evaluating}. Integration of IPCC AR6 into GPT-4 \cite{achiam2023gpt} has advanced conversational AI in climate science\cite{vaghefi2023chatclimate}. \citealt{koldunov2024local} developed a prototype integrating user-relevant information into LLMs for better climate data summarization. ClimateGPT \cite{thulke2024climategpt} introduced climate change domain-specific LLMs optimized for retrieval augmentation and multilingual accessibility.

{\bf VLMs } 
Vision-Language Models (VLMs) like CLIP \cite{radford2021learning}, ALIGN \cite{jia2021scaling}, BLIP\cite{li2022blip}, GLIP \cite{zhang2022glipv2}, and LLaVA \cite{liu2024visual} leverage both techniques of computer vision and natural language processing to reason from visual data. Trained on vast image-text pairs, they excel in tasks like image captioning, visual question answering, object detection, and scene understanding.

{\bf Datasets and Benchmarks for Climate Text }
Several datasets support climate change-related text-based analysis. Climate-fever \cite{diggelmann2020climate} contains 1,535 real-world claims about climate change with annotated evidence from Wikipedia. ClimateBERT-NetZero \cite{schimanski2023climatebert} assesses reduction and net zero emission targets. ClimaText \cite{varini2020climatext} is for climate change topic detection, with labeled sentences from Wikipedia and SEC 10-K filings. CLIMA-INS \cite{spokoyny2023towards} contains survey responses from the NAIC Climate Risk Disclosure Survey, while CLIMA-CDP \cite{spokoyny2023towards} oversees a global disclosure questionnaire with tasks for topic and question classification. ClimateStance \& ClimateEng \cite{vaid2022towards} is a ternary classification dataset from Twitter data for stance detection. SCIDCC \cite{mishra2021neuralnere} contains around 11k news articles with 20 labeled categories relevant to climate change.

As summarized above, existing datasets and benchmarks for LLMs focus on text-based tasks related to climate change. To the best of our knowledge, no dataset or benchmark rigorously examines LLMs' understanding of meteorology concepts. Additionally, no dataset or benchmark evaluates VLM performance in severe weather reasoning using visualizations of current weather conditions. In summary, WeatherQA is the first multimodal dataset for severe weather reasoning, curated from mesoscale analysis, surface observations, and composite radar reflectivity, paired with expert MCDs.

\section{Dataset}
Our goal is to explore the potential of visual language models (VLMs) for understanding and reasoning about severe weather phenomena, specifically thunderstorms and hazardous winter conditions across the CONUS in the upcoming hours (forecast lead-times of several hours). This research aligns with MCDs from the Storm Prediction Center (SPC) \footnote{\url{www.spc.noaa.gov/misc/about.html\#Mesoscale\%20Discussions}} \cite{SPC2024}. We introduce WeatherQA, the first publicly available dataset designed for severe weather reasoning. This dataset features (i) visualizations of ingredients-based forecasting parameters, and (ii) tasks designed to evaluate a VLM's ability to identify affected regions and severe threats. WeatherQA uses data sourced from the NOAA Storm Prediction Center's website,  which is publicly available to access. It pairs MCDs with ingredients-based forecasting parameters at the time of issuance, mirroring the analysis methods used by forecasters to assess weather conditions. The dataset curation procedure is shown in Figure \ref{fig:weatherqa_concept}(A).

\paragraph{Content of Mesoscale Discussions} MCDs include an "Areas affected" line, a "Concerning" line, a valid time, a summary paragraph, a technical discussion paragraph, and a graphical depiction \cite{SPC2024}. We extract and format content based on issue time, area affected, concerning, and summary for each MCD from 2014 to 2020, with an average of 53 words per MCD (e.g., Figure \ref{fig:weatherqa_concept}(A), bottom-left purple box with underscore). Refer to Appendix \ref{appdx: data} for the reasoning behind selecting these years for our dataset. We exclude the graphical product and technical discussion due to data inconsistency across MCDs, which creates confusion for the VLM to learn and understand the meteorological patterns related to different types of potential hazards and their associated regions over the CONUS. Further details on MCD limitations and collection are in Appendix \ref{appdx: Mesoscale Discussions}.

\paragraph{Ingredients-based Forecasting Parameters}\label{Parameters-paragraph} Each MCD is paired with a fixed set of ingredients-based forecasting parameters covering the CONUS, collected at the MCD published time. Each MCD-data pair includes 20 images of common parameters ($800\times600$ pixels), surface observations ($588\times389$ pixels), and composite radar reflectivity ($588\times389$ pixels) from the SPC Hourly Mesoscale Analysis Archive \cite{SPCMA2024}. The composite radar reflectivity shows the highest reflectivity detected from different angles at various heights, providing a clear picture of real-time precipitation. The list of parameters is in Appendix \ref{parameters-appendix}. While snapshots of these parameters may be either excessively or insufficiently detailed, depending on the particular event, they help assess the VLM's ability to visually ground potential severe weather conditions based on ingredients-based parameters. Limitations of these parameters are discussed in Appendix \ref{appdx: data}.

\begin{figure}[t!]
    \centering
    \includegraphics[width=\textwidth]{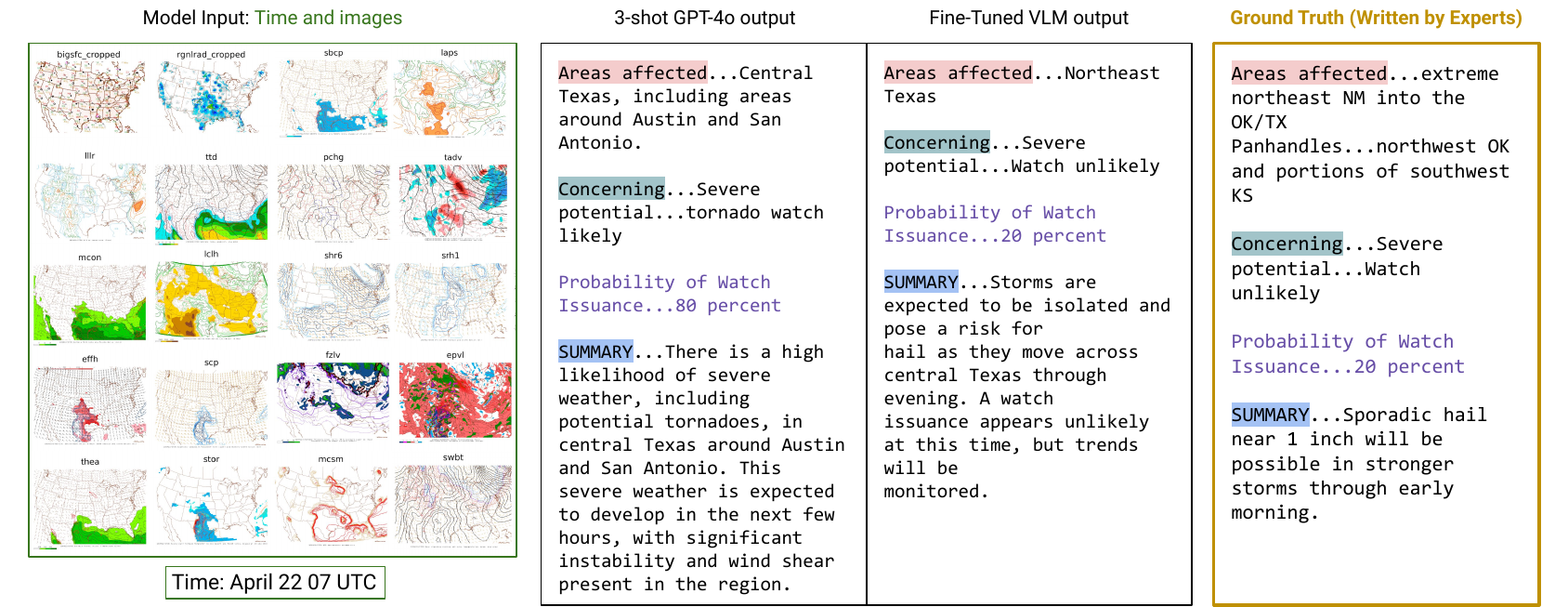}
\caption{Comparison of VLMs' output generated from current timestamp and weather condition images (input) to simplified mesoscale discussions written by experts (ground truth).}
    \label{fig:case study demo}
    \vskip -0.12in
\end{figure}

\paragraph{Dataset Size and Split} WeatherQA comprises a total of 8,511 MCD samples and 170,220 ingredients-based forecasting parameter images collected from 2014 to 2020, with an average of approximately 1,000 samples per year. A breakdown of the types of "Concerning", which serve as classes for the classification task in Section \ref{concerning-classification-task} along with their seasonal counts, is provided in Table \ref{dataset-table}. The category 'Severe potential Watch unlikely' is the most frequently discussed in 'Concerning' across all seasons. The year 2020, being the last year in the dataset, is reserved for the test set. To manage the cost of the experiment and consider the workload for human evaluation, we further downsize the test set to 600 samples for the Multi-Choice QA and classification task, and to 80 samples for the case study, which involves expert evaluation. Since types of severe weather events exhibit seasonality \cite{wang2023climatological} and the number of MCDs varies each month, the test samples are randomly selected based on the monthly distribution of events in 2020.
\section{Tasks}
\subsection{"Areas Affected" Multiple-choice QA}
\paragraph{Task Description} The objective is to test the VLM's capability to identify the geographical area(s) most likely to be affected by severe weather in the near future, based on current weather conditions. An example question and choices are shown in Figure \ref{fig:weatherqa_concept}(B). The correct option is selected from the "Areas affected" line of the chosen MCD for each corresponding test case in the test set, while the three incorrect options are randomly picked from the rest of the dataset. A Jaccard Similarity\footnote{Jaccard Similarity measures the similarity between two sets by dividing the number of common elements by the total number of unique elements in both sets.} \cite{real1996probabilistic} threshold of 0.5 is applied to ensure the incorrect and correct options are sufficiently different.

\paragraph{Evaluation Metrics} We follow settings from other popular Multi-Choice QA benchmarks like HellaSwag \cite{zellers2019hellaswag} and MMLU \cite{hendrycks2020measuring}, which report accuracy as the performance metric. 

\subsection{"Concerning" Multi-class Classification}\label{concerning-classification-task}
\paragraph{Task Description} The objective is to evaluate the VLM's ability to assess the urgency and severity of current weather conditions for a selected area, determine if a severe weather watch is warranted, and identify specific threats related to winter storms, following MCD guidelines \cite{SPC2024}. Typically, MCDs provide detailed meteorological context that informs the probability of issuing a watch or a hazardous winter weather alert when conditions are favorable for development. This information is shown in the "Concerning" line of the MCDs, which uses fixed categories. An example of "Concerning" multi-class classification is shown in Fig \ref{fig:weatherqa_concept}(B) with class names from Table \ref{dataset-table}.

\paragraph{Evaluation Metrics} Due to the class imbalance in the "Concerning" categories of MCDs (Table \ref{dataset-table}), where "severe potential...watch unlikely" is 50\% of the test set, a model that often predicts "watch unlikely" (false negatives) could achieve high accuracy. However, this could underestimate the weather threat or severity, potentially being more harmful than a false positive. Therefore, we report the weighted F1 Score, which calculates the precision and recall for each class and takes a weighted average based on the number of instances per class. This provides a fairer, balanced comparison.

\begin{table}[t!]
  \small
  \caption{Categorization of Weather Concerns by Season in the WeatherQA Dataset}
  \label{dataset-table}
  \centering
  \begin{tabular}{p{6.5cm}p{1cm}p{1cm}p{1cm}p{1cm}p{1cm}}
    \toprule
    \textbf{Weather Concerns} & Winter & Spring & Summer & Autumn & Total/Test\\
    \midrule
    \begin{tabular}[c]{@{}l@{}}Severe potential Watch unlikely\end{tabular} & 269/25 & 1092/84 & 1946/168 & 589/23 & 3896/300\\
    \begin{tabular}[c]{@{}l@{}}Severe potential Watch possible\end{tabular} & 108/3  & 851/48 & 1230/76 & 269/14 & 2458/141\\
    \begin{tabular}[c]{@{}l@{}}Severe potential Watch likely\end{tabular} & 30/5  & 344/25 & 404/23 & 81/0 & 859/53 \\
    \begin{tabular}[c]{@{}l@{}}Heavy snow\end{tabular} &276/25 &95/8 &1/0 &84/11 & 456/44\\
    \begin{tabular}[c]{@{}l@{}}Severe potential tornado watch likely\end{tabular} &40/3 &132/9 &38/5 &32/3 &242/20\\
    \begin{tabular}[c]{@{}l@{}}Severe potential severe thunderstorm watch likely\end{tabular} &3/0 &67/4 &115/14 &6/0 & 191/18\\
    \begin{tabular}[c]{@{}l@{}}Winter mixed precipitation\end{tabular} &151/5 &33/1 &0/0 &10/2 &194/8\\
    \begin{tabular}[c]{@{}l@{}}Freezing rain\end{tabular} &101/8 &14/0 &0/0 &24/3 & 139/11\\
    \begin{tabular}[c]{@{}l@{}}Severe potential watch needed soon\end{tabular} &4/0 &20/1 &20/2 &4/0 &48/3\\
    \begin{tabular}[c]{@{}l@{}}Blizzard\end{tabular} &14/2 &7/0 &0/0 &5/0 &26/2\\
    \begin{tabular}[c]{@{}l@{}}Snow squall\end{tabular} &2/0 &0/0 &0/0 &0/0 &2/0\\
    \midrule
    \begin{tabular}[c]{@{}l@{}}Total/Test\end{tabular} &998/76 &2655/180 &3754/288 &1104/56 &8511/600\\
    \bottomrule
  \end{tabular}
\end{table}

\section{Benchmarks}

\subsection{Baseline Models}

We evaluate state-of-the art VLMs (GPT-4 Turbo \cite{achiam2023gpt}, GPT-4o, Gemini 1.5 Flash,  Gemini 1.5 Pro \cite{reid2024gemini}, Claude3-Opus and Claude 3.5 Sonnet) on our multimodal WeatherQA dataset. We note Claude 3 and 3.5 currently can process a maximum of 20 images, limiting our test to the 0-shot setting. 
All models are configured with a temperature setting of 0.1. Additionally, we designed a Llama3\footnote{https://llama.meta.com/llama3/ } \cite{touvron2023llama} series based VLM and fine-tune it as our baseline. The architecture of our fine-tuned VLM is similar to that of LLaVA \cite{liu2024visual}, with two major modifications: our model accommodates input for 20 images using a shared weights projection layer and encoder for each image, and we employ a parameter-efficient fine-tuning method (LoRA \cite{hu2021lora}) in the second stage of fine-tuning, rather than adjusting the full Llama weights. The whole architecture and training details of our fine-tuned VLM can be found in Appendix \ref{fig:fine-tuned_vlm} and \ref{appdx: fine-tune models}.

\subsection{Prompt Settings}
To evaluate the effectiveness of baseline VLMs on both the MCQ and the multi-class classification task, we experiment with both zero-shot and few-shot prompting strategies. 

\paragraph{Zero-shot Setting}
We evaluate all the VLMs above in zero-shot and a Chain of Thoughts (CoT) zero-shot settings. We partition our prompt template into four parts: \{System Prompt\}, \{Encode Weather Parameters\}, \{Benchmark Prompt Instructions\} and \{Question Template\}. The \{System Prompt\} instructs the VLMs to play the role of an expert in severe weather analysis. {Encode Weather Parameters} includes twenty pairs of ingredient-based weather parameter images with a brief explanation of each parameter and relevant time data. This is designed to provide comprehensive context. To effectively guide the VLMs, the \{Benchmark Prompt Instructions\} specifies three clues for VLMs to follow. For the CoT setting, this section includes step-by-step analysis instructions in the clues within the \{Benchmark Prompt Instructions\} part. Finally, the \{Question Template\} part covers two tasks introduced in this study: The "Areas Affected" Multiple Choice Question and the "Concerning" Multi-class Classification. Further details can be found in Appendix \ref{Function block}.

\paragraph{Few-shot Setting}
We employ 1-shot, 3-shot and 3-shot CoT settings to evaluate the performance of VLMs. Drawing from the zero-shot prompt template, we incorporate N related samples into each few-shot prompt as in-context demonstration. To improve the quality of the few-shot examples, we randomly select the examples from the corresponding month of previous years (e.g., 2018) for each test sample in our test set. Each example consists of \{Encode Weather Parameters\}, \{Question Template\} and the answers for the two questions. See details in Appendix \ref{benchmark prompt}.

\paragraph{Case Study Setting}
In our case study, we use a 3-shot setting for GPT-4o and a zero-shot setting for the fine-tuned VLM. Compared to the objective task, we modified the \{Benchmark Prompt Instructions\} and \{Question Template\}. Instead of multi-choice QA, we instruct the VLMs to identify the precise geographical area that is most likely to be impacted by the potential severe weather. Furthermore, we also instruct the VLMs to generate a 'Summary' section that describes the expected development and evolution of the severe weather (details in Appendix \ref{generation prompt}).

\subsection{Quantitative Results}
\textbf{"Areas Affected" Multi-choice QA}
We show the accuracy of VLMs for a multiple-choice task to identify geographical areas most likely to be affected by severe weather in Table \ref{tab:mcq-accuracy}. Claude 3.5 Sonnet has the highest accuracy with 41.50\% in the 0-shot-CoT setting and achieves 41.17\% in the 0-shot. This is followed by Gemini Pro 1.5 and GPT-4o at 39.00\% and 38.83\% in the 3-shot setting. A breakdown of accuracy by season and storm type is presented in Table \ref{tab: 3-shot GPT-4o MCQ}. Claude3 Opus performs worst, achieving $\sim$20\% in both 0-shot and 0-shot-CoT settings. GPT-4 Turbo, Gemini Flash 1.5, and Fine-tuned Llama3 show lower accuracies (21.33\%-34.33\%). The newer models, such as Claude 3.5 Sonnet, GPT-4o, and Gemini 1.5, demonstrate substantial improvements in accuracy compared to their predecessors, indicating a better contextual understanding of geographical regions in the CONUS with overlaid parameters. Fine-tuned Llama3, still struggles with the localization of weather patterns, highlighting the need for localization in the training or fine-tuning stages. Both Gemini and GPT models show improved accuracy in few-shot settings, but CoT settings yield mixed results: GPT-4o sees improvements while Gemini shows degradation. This indicates difficulties in reasoning about the correlation with their internal knowledge of ingredients-based parameters and geographical regions. Overall, identifying weather-affected areas remains challenging with the latest model Claude 3.5 Sonnet performing roughly 40\%.
\begin{table}[t]
\small
  \caption{Accuracy of Areas Affected Multi-choice QA}
  \label{tab:mcq-accuracy}
  \centering
  \begin{adjustbox}{max width=\textwidth}
  \begin{tabular}{lcccccc}
    \toprule
    & 0-shot & 1-shot & 3-shot & 0-shot-CoT & 3-shot-CoT \\
    \midrule
    Claude 3 Opus & 20.67\% & / & / & 19.17\% & / \\
    Claude 3.5 Sonnet & \textbf{\textcolor{darkgreen}{41.17}}\% & / & / & \textbf{\textcolor{darkgreen}{41.50\%}} & / \\
    GPT-4 Turbo & 21.33\% & 24.17\% & 27.00\% & 23.50\% & 23.33\% \\
    GPT-4o & 36.83\% &  \textbf{\textcolor{darkgreen}{35.67\%}} & 38.83\% & 38.17\% & \textbf{\textcolor{darkgreen}{39.33\%}} \\
    Gemini Flash 1.5 & 30.67\% & 33.00\% & 34.33\% & 31.17\% & 30.67\% \\
    Gemini Pro 1.5 & 31.50\% & \textbf{\textcolor{darkgreen}{35.67\%}} & \textbf{\textcolor{darkgreen}{39.00\%}} & 33.56\% & 33.06\% \\
    \grayrow
    Fine-tuned-VLM (Llama3 8B) & 28.17\% & / & / & / & / \\
    \bottomrule
  \end{tabular}
  \end{adjustbox}
\end{table}

\textbf{"Concerning" Multi-class Classification}
The answer for the "Concerning" classification is provided in the same inference experiments following the `Areas Affected' multiple-choice question. Given the class imbalance with `Severe potential Watch unlikely' accounting for 50\% and `Severe potential Watch possible' accounting for 23.5\% in Table \ref{dataset-table}, it is crucial to consider class distribution when evaluating performance. In the 0-shot setting, the Fine-tuned Llama3 achieves the highest accuracy (45\%) and weighted F1 score (42\%). However, considering the class imbalance, an accuracy of 45\% is not as impressive as it seems. Among the proprietary models, GPT-4o outperforms others in terms of weighted F1 scores achieving 0.31 in the 3-shot setting. Gemini Pro 1.5 closely follows with a weighted F1=0.21 in both 1-shot and 3-shot settings. Note that the Claude 3.5 Sonnet achieves an accuracy of 18\% which is a substantial improvement from Claude 3 Opus and outperforms the rest of the proprietary models by a large margin. The weighted F1 scores for all models are low compared to their accuracy scores in zero-shot scenarios, indicating that the models struggle to correctly classify the minority classes. The few-shot settings generally improve performance, suggesting that examples help the models understand the concerns and handle class imbalance. However, the CoT settings show mixed results with GPT-4o and Gemini models. Future work should improve alignment between geographical regions to weather patterns and improve the models' ability to classify minority classes in MCDs.

\begin{table}[t]
\normalsize
  \caption{Accuracy / Weighted F1 Score of Concerning Classification }
  \label{tab:classification-accuracy}
  \centering
  \begin{adjustbox}{max width=\textwidth}
  \begin{tabular}{lcccccc}
    \toprule
    & 0-shot & 1-shot & 3-shot & 0-shot-CoT & 3-shot-CoT \\
    \midrule
    Claude 3 Opus & 3.50\% / 0.01 & / & / & 3.33\% / 0.01 & / \\
    Claude 3.5 Sonnet & 13.50\% / 0.1 & / & / & \textbf{\textcolor{darkgreen}{18.00\%}} / \textbf{\textcolor{darkgreen}{0.1}} & / \\
    GPT4-Turbo & 3.67\% / 0.04 & 12.50\% / 0.09 & 5.50\% / 0.06 & 3.00\% / 0.01 & 5.33\% / 0.04 \\
    GPT4-o & 8.17\% / 0.03 & 22.33\% / \textbf{\textcolor{darkgreen}{0.21}} & \textbf{\textcolor{darkgreen}{28.83\%}} / \textbf{\textcolor{darkgreen}{0.31}} & 8.00\% / 0.05 & 8.33\% / 0.05 \\
    Gemini Flash 1.5 & 7.17\% / 0.02 & 17.67\% / 0.19 & 13.00\% / 0.14 & 3.33\% / 0.01 & 5.83\% / 0.04 \\
    Gemini Pro 1.5 & 4.67\% / 0.02 & \textbf{\textcolor{darkgreen}{25.33\%}} / \textbf{\textcolor{darkgreen}{0.21}} & 24.17\% / 0.21 & 2.67\% / 0.01 & \textbf{\textcolor{darkgreen}{18.70\%}} / \textbf{\textcolor{darkgreen}{0.17}} \\
    \grayrow
    Fine-tuned-VLM (Llama3 8B)  & \textbf{\textcolor{darkgreen}{45.00\%}} / \textbf{\textcolor{darkgreen}{0.42}} & / & / & / & / \\
    \bottomrule
  \end{tabular}
  \end{adjustbox}
\end{table}

\section{Case Study: Mesoscale Discussion Generation}
We present a case study on generating simplified MCDs with GPT-4o and finetuned-VLM Llama2 \cite{touvron2023llama2}, emulating forecasters' workflows. The VLMs generate the "Areas Affected," "Concerning," and "Summary" lines. We evaluate their ability to forecast threat timing, coverage, intensity, and type. Examples are shown in Fig \ref{fig:case study demo} (prompts in Appendix \ref{generation prompt}). Some samples lack surface observations or radar reflectivity (e.g. Figure \ref{2020MCD1833}), reflecting real forecast scenarios. We use BLEU \cite{papineni2002bleu}, BERTScore \cite{zhang2019bertscore}, ROUGE-Lsum \cite{lin2004rouge}, and METEOR \cite{banerjee2005meteor}, and calculate Spearman correlation \cite{zar2005spearman} against human expert ratings. This provides further insight on if current metrics can evaluate VLM weather reasoning.

\paragraph{The Need for Expert Evaluation}
 VLM outputs structurally resemble actual MCDs their details differ. This highlights the need for nuanced evaluation beyond metrics like accuracy and F1 score. To this end, we conduct a comprehensive evaluation with experts forecasters. They are tasked with assessing the quality, accuracy, and relevance of VLM outputs, ensuring they provide valuable insights and adhere to MCD conventions. Four active forecasting experts graded and evaluated outputs of GPT-4o and Llama2, in a thorough review process (see Appendix \ref{expert-review-procedure}).

\paragraph{Grading Rubric and Insights}
A tailored grading rubric (see Appendix \ref{rubrics}) evaluates: 1) accuracy of geographical areas, 2) correctness of severe weather potential and watch issuance probability, and 3) rationality of the summary. Scores range from 0 to 9, with overall performance categorized as poor (0-3), fair (4-6), good (7-8), or excellent (9). We will release the grades for each sample. Guidelines with feedback (Appendix \ref{Guidelines}) will help future researchers assess VLM-generated outputs, identifying strengths and weaknesses to improve VLM performance in weather forecasting.

\subsection{Ratings}

\begin{table}[h]
\vspace{-10pt}
\caption{Comparison of generations from VLMs. BLEU, BERTScore, ROUGE-Lsum, and METEOR columns show the correlation of these metrics with expert ratings. The distribution of expert ratings is illustrated in Figure \ref{fig: expert ratings}.}
  \label{tab:case-study-metrics}
  \centering
  \begin{adjustbox}{max width=\textwidth}
  \begin{tabular}{l|c|cccc}
    \toprule
    & Expert ratings & BLEU & BERTScore & ROUGE-Lsum & METEOR  \\
    \midrule
    3-shot GPT-4o & \textbf{\textcolor{darkgreen}{2.288}} & 0.138 / \textbf{\textcolor{darkgreen}{0.403}} & 0.864 / \textbf{\textcolor{darkgreen}{0.355}} & 0.355 / \textbf{\textcolor{darkgreen}{0.371}} & 0.264 / \textbf{\textcolor{darkgreen}{0.294}} \\
    \grayrow
    Fine-tuned-VLM Llama2 & 2.213 & \textbf{\textcolor{darkgreen}{0.267}} / 0.286 & \textbf{\textcolor{darkgreen}{0.899}} / 0.284 & \textbf{\textcolor{darkgreen}{0.388}} / 0.321 & \textbf{\textcolor{darkgreen}{0.426}} / 0.272 \\
    \bottomrule
  \end{tabular}
  \end{adjustbox}
\end{table}

The 3-shot GPT-4o performs similarly to Fine-tuned VLM Llama2 as shown in Table \ref{tab:case-study-metrics}, and both have a rating of 2.2 which is categorized as "Poor Performance" according to the Grading Rubric in Appendix \ref{rubrics}. Although the fine-tuned VLM Llama2 achieves relatively higher scores in typical evaluation metrics, the lower correlation score compared to 3-shot GPT-4o suggests that the fine-tuned VLM Llama2 mainly gains knowledge in the format regarding how the MCDs are written rather than the essence or intention of weather reasoning presented in MCDs. Both models' performance not only suggests the need for further research in text summarization metrics but also highlights a substantial knowledge gap in VLMs when it comes to providing insights into weather reasoning.

\subsection{Expert Feedback}\label{expert_feedbacks}
The VLMs showed mixed performance which impacted forecasters' confidence and trust. While GPT-4o, addressed similar hazards and timing, it had issues such as spatial displacements, incorrect hazard identification, and incorrectestimation of severe weather potential. Despite this, experts recognized the potential of VLMs to aid forecasters. The examples below showcase VLM limitations:
\begin{itemize}[leftmargin=20pt]
    \item VLMs showed mixed performance in accuracy and reliability, affecting forecasters' confidence. For instance, GPT-4o in 2020MCD1833 (Appendix Figure \ref{2020MCD1833}) had good overlap with actual hazards and timing, but significant spatial displacements and incorrect hazard identification were noted, such as in 2020MCD1611 (Appendix Figure \ref{2020MCD1611}). 
    \item VLMs varied in specificity, sometimes providing accurate hazard information but also failing to specify main threats or giving overly broad reasonings, such as when GPT-4o inappropriately predicted tornadoes in 2020MCD1611. Experts further noted that VLMs often misinterpreted weather maps, focusing on specific parameters without broader context.
    \item Instances of hallucinations and inaccuracies were observed, such as Fine-tuned VLM's unreasonable probabilities and nonsensical language in 2020MCD0061 (Appendix Figure \ref{2020MCD0061}). VLMs showed inconsistent performance and limited contextual understanding, sometimes aligning with actual MCDs but also targeting irrelevant atmospheric regimes.
    \item Language clarity varied, with some discussions being coherent and others confusing. For example, in 2020MCD0303 (Appendix Figure \ref{2020MCD0303}), GPT-4o's use of `moderate risk' was inappropriate and could cause public confusion.
    \item VLMs seem to be mostly chasing convective available energy and shear as reflected in 2020MCD0303. Areas for improvement include addressing spatial displacements, over-reliance on certain parameters, and generation of false alarms.
\end{itemize}
 Experts also highlighted the potential strength of VLMs for assisting weather forecasters. As one expert noted: "Fine-tuned VLM was pretty good in placement relative to 2020MCD565 (Figure \ref{2020MCD0565}). It got the probability of watch issuance right, and even explicitly listed severe winds as a threat, where the forecaster did not specify the main threat associated with mesoscale convective system. Lapse rates, forcing, and shear were all present here. Only one hail report was received in this area, so both the performance of the model and the forecaster are comparable." They suggested that with further refinement, these models could offer valuable insights to assist forecasters in drafting MCDs.

\section{Conclusion}
WeatherQA is the first multimodal dataset designed to advance models' ability to reason about severe weather using ingredients-based weather parameters, surface observations, and radar reflectivity, paired with expert-written weather discussions. It presents two tasks with benchmarks from state-of-the-art VLMs: analyzing the potentially affected areas through multi-choice QA and determining the development potential of severe convection via classification, requiring geographical knowledge, weather forecasting practices, and complex weather domain expertise.

Expert feedback on VLM generated weather summaries reveals significant gaps in reasoning, including hallucinations about event severity and location. Future work should improve the understanding of atmospheric dynamics context for VLMs and curate datasets with better geographical localization of severe weather to enhance the quality and reliability of VLMs for weather reasoning.

\begin{ack}
ZH and AAF are supported by NSF Award Number 2209699. The authors thank Storm Prediction Center forecasters Bryan Smith, Andrew Moore, Brian Squitieri, and Matt Mosier for their expert evaluation of the model.

\end{ack}

\bibliography{neurips_data_2024}


\section*{Checklist}

\begin{enumerate}

\item For all authors...
\begin{enumerate}
  \item Do the main claims made in the abstract and introduction accurately reflect the paper's contributions and scope?
    \answerYes{}
  \item Did you describe the limitations of your work?
    \answerYes{} The limitation is discussed in Appendix \ref{appdx: data}
  \item Did you discuss any potential negative societal impacts of your work?
    \answerYes{} The potential negative societal impacts are mentioned in Appendix \ref{Impact statement}.
  \item Have you read the ethics review guidelines and ensured that your paper conforms to them?
    \answerYes{}
\end{enumerate}

\item If you are including theoretical results...
\begin{enumerate}
  \item Did you state the full set of assumptions of all theoretical results?
    \answerNA{}
	\item Did you include complete proofs of all theoretical results?
    \answerNA{}
\end{enumerate}

\item If you ran experiments (e.g. for benchmarks)...
\begin{enumerate}
  \item Did you include the code, data, and instructions needed to reproduce the main experimental results (either in the supplemental material or as a URL)?
    \answerYes{} The URL to the code, data, and instructions will be provided in the supplemental material
  \item Did you specify all the training details (e.g., data splits, hyperparameters, how they were chosen)?
    \answerYes{} See Appendix \ref{appdx: model}.
	\item Did you report error bars (e.g., with respect to the random seed after running experiments multiple times)?
    \answerNo{} The test size is sufficiently large to reliably assess the performance of different models without the need for error bars.
	\item Did you include the total amount of compute and the type of resources used (e.g., type of GPUs, internal cluster, or cloud provider)?
    \answerYes{} See Appendix \ref{appdx: model}.
\end{enumerate}

\item If you are using existing assets (e.g., code, data, models) or curating/releasing new assets...
\begin{enumerate}
  \item If your work uses existing assets, did you cite the creators?
    \answerYes{}
  \item Did you mention the license of the assets?
    \answerYes{} See dataset URL in supplemental material
  \item Did you include any new assets either in the supplemental material or as a URL?
    \answerYes{}
  \item Did you discuss whether and how consent was obtained from people whose data you're using/curating?
    \answerYes{} We follow the copyright of the original data source from National Oceanic and Atmospheric Administration (NOAA)
  \item Did you discuss whether the data you are using/curating contains personally identifiable information or offensive content?
    \answerNA{}
\end{enumerate}

\item If you used crowdsourcing or conducted research with human subjects...
\begin{enumerate}
  \item Did you include the full text of instructions given to participants and screenshots, if applicable?
    \answerNA{}
  \item Did you describe any potential participant risks, with links to Institutional Review Board (IRB) approvals, if applicable?
    \answerNA{}
  \item Did you include the estimated hourly wage paid to participants and the total amount spent on participant compensation?
    \answerNA{}
\end{enumerate}

\end{enumerate}


\clearpage

\appendix
\input{appendix_impact.tex}
\input{appendix_data.tex}
\input{appendix_model.tex}
\input{appendix_prompt.tex}
\input{appendix_case_study.tex}

\clearpage
\newpage
\input{supplementary.tex}

\end{document}

%% file: appendix_impact.tex
\section{Impact Statement}\label{Impact statement}
 While the purpose of proposing WeatherQA is to arouse the community's interest in improving weather reasoning for multimodal models which could help mitigate risks to human life and infrastructure, we recognize potential negative impacts that warrant discussion. First, if the models exhibit geographic or demographic biases in their analysis and predictions, this could lead to inequitable allocation of false alarms or misses across regions or populations. Rigorous bias testing and mitigation strategies would be needed. Second, issuing mesoscale discussions, convective outlook, watches, and warnings for severe weather events requires extremely reliable models; if model hallucinations lead to missed events, the consequences could be severe. Model uncertainty quantification and human oversight are critical for any attempt to use multimodal models for weather analysis and prediction. Stating these concerns as part of the WeatherQA release is intended to promote positive societal impacts while mitigating the risk of abusing multimodal models for weather-related tasks.

%% file: appendix_data.tex
\section{Data}\label{appdx: data}
\subsection{Ingredients-based Forecasting Parameters}\label{parameters-appendix}
Table \ref{parameters-table} lists the ingredients-based forecasting parameters curated from the Mesoscale Analysis Archive. These are the common ingredient-based composite parameters that describe whether the environment is favorable for deep moist convection. The composite parameters allow forecasters to quickly diagnose potential regions for convection based on typical thresholds determined for each parameter. As previously mentioned in section \ref{Parameters-paragraph}, a favorable condition for convection does not guarantee the convection initiation or occurrence of severe weather. Extra information from other weather parameters may be required to further identify the probability and location of severe weather.
\begin{table}[h]
  \small
  \caption{Selected Ingredients-based Forecasting Parameters and their abbreviations in the Mesoscale Analysis Archive}
  \label{parameters-table}
  \centering
  \begin{tabular}{ll}
    \toprule
    Abbreviation & Parameter Name \\
    \midrule
    bigsfc & Regional Surface Chart (surface observation plots)\\
    rgnlrad & Base Reflectivity Mosaic (composite radar reflectivity) \\
    sbcp & Surface-based Convective Available Potential Energy and Convective Inhibition\\
    laps & 700-500mb lapse rate instability \\
    lllr & 0-3km lapse rate instability \\
    ttd & Surface temperature, dewpoint, and Mean Sea Level pressure \\
    thea & Surface equivalent potential temperature advection \\
    pchg & Two-hour surface pressure change \\
    tadv & 850 mb Temperature Advection \\
    mcon & Deep layer moisture flux convergence \\
    lclh & Lifting Condensation Level height \\
    shr6 & 0-6km wind shear magnitude \\
    srh1 & 0-1km storm-relative helicity \\
    effh & Storm-relative helicity in the effective layer \\
    scp & Supercell Composite Parameter \\
    stor & Significant Tornado Parameter (fixed layer) \\
    mcsm & Mesoscale Convective System maintenance probability \\
    fzlv & Freezing Level Info \\
    epvl & 800-750mb Geostrophic Equivalent Potential Vorticity \\
    swbt & Surface wet-bulb temperature \\
    \bottomrule
  \end{tabular}
\end{table}

\subsection{Mesoscale Analysis Archive}
The link to Mesoscale Analysis Archive is located at\url{https://www.spc.noaa.gov/exper/ma_archive/}. This archive stores hourly snapshots of mesoscale analysis images across a wide range of weather parameters, starting from 2005. However, the archive does not guarantee the presence of all weather parameters, which may result in blank or repeated images. We created a filtered dataset that includes all available parameters described in section \ref{parameters-appendix}. WeatherQA only extends up to 2020 because the Regional Surface Chart and Base Reflectivity Mosaic are not available in the Mesoscale Analysis Archive after that year. We also provide an example case study with an empty regional surface chart and a base reflectivity mosaic in Appendix \ref{cases-appendix} Figure \ref{2020MCD1833}.

\subsection{Limitations}
The current curated dataset contains only one static image of each parameter in the mesoanalysis based on the time of issuance of the Mesoscale Discussions (MCDs). Ideally, the evolution of these parameters, as well as the ambient mesoscale or synoptic environments over time, could be just as important in determining the specifics of an MCD. Certain parameters that were not included, such as upper-air plots (e.g., 850 MB or 500 mb geopotential height), could also be crucial for understanding how strong the forcing would be to support convection. Nonetheless, we have decided to focus on the ingredient-based parameters, as they provide a reasonable first guess of the potential for severe weather. Current benchmarks have already shown that VLMs struggle even with these ingredient-based analyses (or pattern matching). The goal is to arouse the community's interest in weather reasoning for VLMs and to convey that reports like MCDs and Watches are valuable resources for LLM training and fine-tuning in the future.

\subsection{Mesoscale Discussions (MCDs)}\label{appdx: Mesoscale Discussions}
A sample page within the Mesoscale Discussions Archive is \url{https://www.spc.noaa.gov/products/md/2020/}. The year can be changed by replacing the last four digits of the URL. The MCDs can be downloaded as files that end with the suffix \textit{.txt}. MCDs with \textit{.txt} are only available after 2014, which is the start year of WeatherQA. We further filtered out MCDs that contain specific 'Severe thunderstorm watch' or 'Tornado watch' alerts, as those MCDs typically describe already ongoing severe weather. Nevertheless, these scenarios could potentially serve as another dataset for understanding the maintenance of severe convective storms.
\paragraph{Potential and Limitations}\label{MCDs-limitation}
Currently, only part of the MCDs are utilized in creating WeatherQA. The 'DISCUSSION' section, which contains detailed technical analysis of the current weather conditions and speculation about future weather, is not included in WeatherQA to avoid overcomplicating the initial evaluation of results. However, there is great potential to help VLMs gain a deeper understanding of meteorology by including such technical analysis, coupled with intelligent sampling of weather conditions, to train or fine-tune VLMs. Geolocation information within the graphics product of the MCD is also a valuable resource for helping VLMs learn locations mapped by forecasters. The chances of misinterpreting weather conditions could be alleviated with better matches in area prediction.

%% file: appendix_model.tex
\section{Model}\label{appdx: model}
\subsection{Training dataset}
We utilize samples from 2014 to 2019 as our training dataset, which comprises 7,331 MCD text samples and 146,620 image samples of ingredient-based parameters. To construct the training prompts, we adhere to the same methodology outlined in the zero-shot and case study settings. For more detailed information, please refer to Appendices \ref{Function block} and \ref{generation prompt}.
\subsection{Training Procedure}\label{appdx: fine-tune models}
For the training of our fine-tuned VLM model, we consider a three-stage procedure. Firstly, to encode the image of ingredients-based parameters effectively, we utilize the masked autoencoder\cite{he2022masked} method to train an encoder with the 146,260 image samples of ingredients-based parameters. In the second stage, we do the pre-training for feature alignment. Specifically, we keep the LLM weights frozen, and update both the weights of the pre-trained encoder and the projection layer. This stage can be understood as aligning the ingredients-based parameters feature with the pre-trained LLM word embedding. For the third stage, we continue to update both the weights of the pre-trained encoder and the projection layer and employ a parameter-efficient fine-tuning method (LoRA) to fine-tune the LLM. We perform instruction-tuning of the LLM on the prediction tokens in both stage 2 and 3, using its original auto-regressive training objective. 

Our VLM is trained using two Nvidia L40s, following the Llama2 7B and Llama3 8B's hyperparameters. Due to time constraints, we use the Llama2 7B as the LLM backbone for the case study and the Llama3 8B for the objective tasks. In stage 1, the masked autoencoder is trained with a learning rate of 1e-3 and a batch size of 128 over 25 epochs(about 10 hours). In stage 2, the VLM is trained with a learning rate of 5e-5, a batch size of 4 and 20 epochs(about 12 hours). In stage 3, the VLM is trained with a learning rate of 2e-5, a batch size of 2 and 15 epochs(about 15 hours). The LoRA \cite{hu2021lora} settings for this stage include a rank of 8 and an alpha of 16, focusing on weights type \(W_q\), \(W_k\) and \(W_v\). 

\begin{figure}
    \centering
    \includegraphics[width=\textwidth]{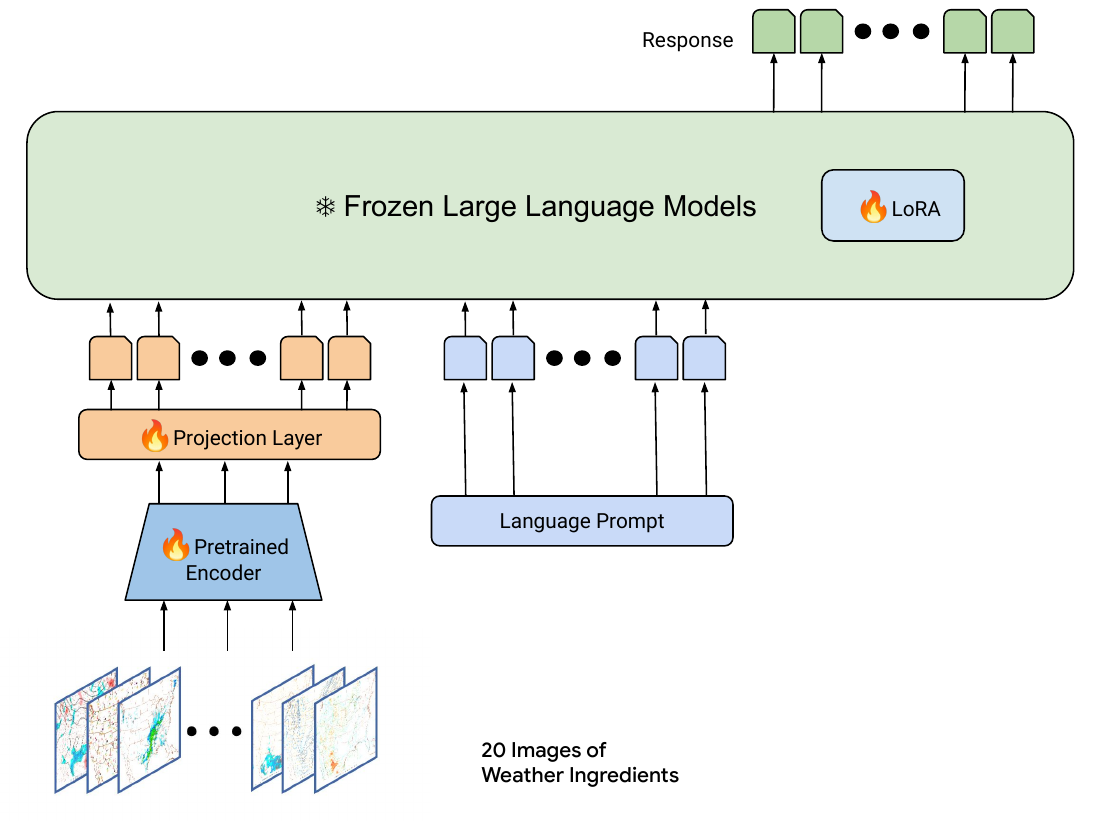}
    \caption{\textbf{An overview of Fine-tuned VLM architecture} }
    \label{fig:fine-tuned_vlm}
\end{figure}

\subsection{Extra analysis of the benchmark}

\begin{table}[ht]
  \caption{3-shot GPT-4o Area Affected multiple-choice performance breakdown by season and concerning type for winter versus non-winter severe concerns}
  \label{tab: 3-shot GPT-4o MCQ}
  \centering
  \begin{tabular}{lcccccc}
    \toprule
    \multicolumn{1}{c}{} & \multicolumn{2}{c}{Winter Severe} & \multicolumn{2}{c}{Non-Winter Severe} \\
    \cmidrule(r){2-3} \cmidrule(r){4-5}
    & acc & sample count & acc & sample count \\
    \midrule
    Winter & 0.55 & 40 & 0.5 & 36 \\
    Spring & 0.333 & 9 & 0.404 & 171 \\
    Summer & - & - & 0.316 & 288 \\
    Autumn & 0.438 & 16 & 0.575 & 40 \\
    \bottomrule
  \end{tabular}
\end{table}

\begin{table}[ht]
  \caption{3-shot GPT-4o Concerning classification performance breakdown by season and concerning type for winter versus non-winter severe concerns}
  \label{tab: 3-shot GPT-4o Classification}
  \centering
  \begin{tabular}{lcccc}
    \toprule
    \multicolumn{1}{c}{} & \multicolumn{2}{c}{Winter Severe} & \multicolumn{2}{c}{Non-Winter Severe} \\
    \cmidrule(r){2-3} \cmidrule(r){4-5}
    & acc & weighted F1 & acc & weighted F1 \\
    \midrule
    Winter & 0.2 & 0.286 & 0.361 & 0.478 \\
    Spring & 0.0 & 0.0 & 0.251 & 0.267 \\
    Summer & - & - & 0.333 & 0.338 \\
    Autumn & 0.063 & 0.115 & 0.3 & 0.385 \\
    \bottomrule
  \end{tabular}
\end{table}

3-shot GPT-4o is selected for further analysis as it has relatively good performance in both the multiple-choice task and the classification task. We split the test results by season and further split the "Concerning" classes into two categories. "Winter Severe" includes 'Heavy snow', 'Winter mixed precipitation', 'Freezing rain', 'Blizzard' and 'Snow squall'. 'Non-winter Severe' includes the rest of the categories related to severe potential watches.

In the multiple-choice task (Table \ref{tab: 3-shot GPT-4o MCQ}), 3-shot GPT-4o performs best in winter for winter severe concerns with 0.55 accuracy and in autumn for non-winter severe concerns with 0.575 accuracy. The lowest performance is observed in summer for non-winter severe, which has the largest number of samples, with 0.316 accuracy.

For the classification task (Table \ref{tab: 3-shot GPT-4o Classification}), 3-shot GPT-4o's accuracy and weighted F1 scores are generally lower. The highest accuracy for winter severe concerns is in winter with 0.2 accuracy and 0.286 weighted F1. As for non-winter severe concerns, winter also shows the highest performance but with fewer samples. Spring shows the poorest performance for non-winter severe classees.

%% file: appendix_prompt.tex
\section{Prompt}

\subsection{Function block for Benchmark}\label{Function block}

\begin{tcolorbox}[colback=yellow!10!white, colframe=yellow!50!black, title=<System Prompt>]\label{system prompt}
As an AI assistant with expertise in severe weather analysis and forecasting, you are equipped to interpret a comprehensive figures that illustrating various weather variables crucial for understanding the latest weather conditions across the contiguous United States. Your responsibility as a weather forecaster is to provide accurate and timely insights into weather conditions and potential severe weather threats.
\end{tcolorbox}

\begin{tcolorbox}[colback=white, colframe=black, title= <Encode Weather Parameters>, breakable]\label{encode weather parameters}
\begin{itemize}
    \item \textbf{Regional Surface Chart}: Represents regional surface weather conditions.
    \item \textbf{Base Reflectivity Mosaic}: Shows the base reflectivity from radar observations.
    \item \textbf{Surface-based Convective Available Potential Energy (CAPE) and Convective Inhibition (CIN)}: Indicates the potential energy available for convection and the inhibition preventing convection.
    \item \textbf{700-500mb Lapse Rate Instability}: Displays the instability in the 700-500mb layer of the atmosphere.
    \item \textbf{0-3km Lapse Rate Instability}: Indicates the instability in the 0-3km layer of the atmosphere.
    \item \textbf{Surface Temperature, Dewpoint, and Mean Sea Level (MSL) Pressure}: Shows the surface temperature, dewpoint, and MSL pressure.
    \item \textbf{Surface Equivalent Potential Temperature Advection}: Displays the advection of equivalent potential temperature at the surface.
    \item \textbf{Two-hour Surface Pressure Change}: Shows the change in surface pressure over two hours.
    \item \textbf{850 mb Temperature Advection}: Indicates the advection of temperature at the 850 mb level.
    \item \textbf{Deep Layer Moisture Flux Convergence}: Displays the moisture flux convergence in the deep layer.
    \item \textbf{Lifting Condensation Level Height}: Shows the height of the lifting condensation level.
    \item \textbf{0-6km Wind Shear Magnitude}: Indicates the magnitude of wind shear in the 0-6km layer.
    \item \textbf{0-1km Storm-relative Helicity}: Displays the storm-relative helicity in the 0-1km layer.
    \item \textbf{Storm-relative Helicity in the Effective Layer}: Indicates the helicity in the effective layer.
    \item \textbf{Supercell Composite Parameter}: Shows the supercell composite parameter.
    \item \textbf{Significant Tornado Parameter (fixed layer)}: Displays the significant tornado parameter for a fixed layer.
    \item \textbf{Mesoscale Convective System Maintenance Probability}: Indicates the probability of mesoscale convective system maintenance.
    \item \textbf{Freezing Level}: Shows the freezing level in the atmosphere.
    \item \textbf{Geostrophic Equivalent Potential Vorticity}: Displays the geostrophic equivalent potential vorticity.
    \item \textbf{Surface Wet-bulb Temperature}: Indicates the surface wet-bulb temperature.
\end{itemize}

\textbf{Repeat for each parameter above and append to <Encoded Parameters>:}

\hspace{1em}\hangindent=1em\hangafter=1 Encode the image parameter by appending its parameter description and its corresponding encoded image.\newline

\textbf{<Encode Weather Parameters>} output: The following 20 figures represent weather conditions at \textbf{[Time of ingredients-based parameters]} and each figure contains multiple weather parameters, the most important variable in each figure is provided in \textbf{<Encoded Parameters>}

\end{tcolorbox}

\begin{tcolorbox}[colback=blue!10!white, colframe=blue!50!black, title=<Benchmark Prompt Instructions>, label=Benchmark_Prompt_Instructions]
Use the following clues to answer the following multiple-choice and classification questions:
\begin{enumerate}
    \item Go through each depicted weather field within the figures in \textbf{<Encode Weather Parameters>} for weather analysis one by one and consider whether each weather analysis field shows signs of potential for severe weather.
    \item Combine the relevant weather fields that have the potential for severe weather and assess their severity to reason out the answer to the question.
    \item \textbf{If not CoT:} Map the answer to one of the options. Only give the correct option character, without any other details or explanations. Output in JSON with the keys "Area\_Affected" for the first answer and "Concerning" for the second answer. \textbf{Else:} Answer both questions in \textbf{<Problem>} after providing a step-by-step analysis. Output in JSON with the key \texttt{"Analysis"} for step-by-step analysis and the final answer to the question with the keys \texttt{"Area\_Affected"} for the first answer and \texttt{"Concerning"} for the second answer with only the correct option character.
\end{enumerate}
\end{tcolorbox}

\begin{tcolorbox}[colback=green!10!white, colframe=green!50!black, title=<Question Template>]\label{Question Template}
\textbf{Areas Affected:} Choose one geographical area(s) most likely to be impacted by the severe weather event from the four options provided.\newline

\hspace*{\fill}Example \textbf{[MCQ options]} provided for demonstration:\hspace*{\fill}\newline

\hspace{3em}A: "South-Central NE"\newline
\hspace{4em}\hangindent=3em\hangafter=1 B: "Portions of southwest and south-central South Dakota"\newline
\hspace{4em}\hangindent=3em\hangafter=1 C: "Portions of central Louisiana into Mississippi"\newline
\hspace{4em}\hangindent=3em\hangafter=1 D: "Northern Arkansas and adjacent southern Missouri"\newline

\textbf{Concerning:} Choose the most likely scenario regarding severe weather concerns from the options provided. Options include whether a convective watch has been issued, the probability of a future watch, or, in winter, potential weather phenomena related to winter storms. \newline

\hspace*{\fill}The options are the types of "Concerning" presented in Table \ref{dataset-table}.\hspace*{\fill}

\end{tcolorbox}

\subsection{Benchmark Prompt}\label{benchmark prompt}

\begin{tcolorbox}[colback=white, colframe=red, title=Benchmark Prompt Demonstration based on function block \ref{Function block}, breakable]
First, configure the system prompt based on \textbf{<System prompt>} and start with \textbf{<Benchmark Prompt Instructions>}.\newline
Second, configure function block \textbf{<Benchmark Prompt Instructions>} to CoT or No-CoT.\newline
Next, select \textbf{<Few-shot>} if needed. \newline
Finally, append the \textbf{<Problem>}. 
\end{tcolorbox}

\begin{tcolorbox}[colback=red!10!white, colframe=red!50!black, title=<Few-shot>]
\textbf{Repeat N times for few-shot:}

\hspace{1em}\texttt{Example ith input:}

\hspace{3em}\hangindent=3em\hangafter=1 Modify \textbf{<Encode Weather Parameters>} given \textbf{[Time of ingredients-based parameters]} and \textbf{<Encoded parameters>}, and \textbf{[MCQ options]} for Area Affected in \textbf{<Question Template>}

\hspace{1em}\texttt{Example ith answer:}
\begin{itemize}
    \item <Area affected answer>{correct option}</Area affected answer>
    \item <Concerning answer>{correct option}</Concerning answer>
\end{itemize}

\end{tcolorbox}

\begin{tcolorbox}[colback=gray!10!white, colframe=gray!50!black, title=<Problem>]
Modify \textbf{<Encode Weather Parameters>} given \textbf{[Time of ingredients-based parameters]} and \textbf{<Encoded parameters>}, and \textbf{[MCQ options]} for Area Affected in \textbf{<Question Template>}

\hspace{1em}\texttt{Output format:}
\begin{itemize}
    \item <Area affected answer>{option}</Area affected answer>
    \item <Concerning answer>{option}</Concerning answer>
\end{itemize}
\end{tcolorbox}

\subsection{Function block for Generation}\label{function block generation}

\begin{tcolorbox}[colback=blue!10!white, colframe=blue!50!black, title=<Generation Prompt Instructions>, label=Benchmark_GenPrompt_Instructions]
Use the following clues to answer the following multiple-choice and classification questions:
\begin{enumerate}
    \item Go through each depicted weather field within the figures in \textbf{<Encode Weather Parameters>}\ref{Function block} for weather analysis one by one and consider whether each weather analysis field shows signs of potential for severe weather.
    \item Combine the relevant weather fields that have the potential for severe weather and assess their severity to reason out the answer to the question.
    \item Provide the potential severe weather analysis report based on the relevant weather fields, and adhere to each question's guidelines strictly.
    \item Use JSON format with the keys "Area Affected", "Concerning", and "Summary" for each of the following questions, respectively.
\end{enumerate}
\end{tcolorbox}

\begin{tcolorbox}[colback=green!10!white, colframe=green!50!black, title=<Generation Question Template>]
\textbf{Areas Affected:} Provide a single concise sentence that describes the precise geographical area(s) most likely to be impacted by the potential severe weather. Avoid referring to broad and vague geographical regions.

\textbf{Concerning:} Select the most appropriate phrase that describes the type and probability of watch or hazardous winter weather events from the following options and output its content without the option alphabet: 

\hspace{3em}A: "Severe potential...Watch unlikely   Probability of Watch Issuance...5 percent"\newline
\hspace{4em}\hangindent=3em\hangafter=1 B: "Severe potential...Watch unlikely   Probability of Watch Issuance...20 percent"\newline
\hspace{4em}\hangindent=3em\hangafter=1 C: "Severe potential...Watch possible   Probability of Watch Issuance...40 percent"\newline
\hspace{4em}\hangindent=3em\hangafter=1 D: "Severe potential...Watch possible   Probability of Watch Issuance...60 percent"\newline
\hspace{4em}\hangindent=3em\hangafter=1 E: "Severe potential...Watch likely   Probability of Watch Issuance...80 percent"\newline
\hspace{4em}\hangindent=3em\hangafter=1 F: "Severe potential...Watch likely   Probability of Watch Issuance...95 percent"\newline
\hspace{4em}\hangindent=3em\hangafter=1 G: "Severe potential...tornado watch likely   Probability of Watch Issuance...80 percent"\newline
\hspace{4em}\hangindent=3em\hangafter=1 H: "Severe potential...tornado watch likely   Probability of Watch Issuance...95 percent"\newline
\hspace{4em}\hangindent=3em\hangafter=1 I: "Severe potential...severe thunderstorm watch likely   Probability of Watch Issuance...80 percent"\newline
\hspace{4em}\hangindent=3em\hangafter=1 J: "Severe potential...severe thunderstorm watch likely   Probability of Watch Issuance...95 percent"\newline
\hspace{4em}\hangindent=3em\hangafter=1 K: "Severe potential...watch needed soon   Probability of Watch Issuance...95 percent"\newline
\hspace{4em}\hangindent=3em\hangafter=1 L: "Heavy snow"\newline
\hspace{4em}\hangindent=3em\hangafter=1 M: "Winter mixed precipitation"\newline
\hspace{4em}\hangindent=3em\hangafter=1 N: "Freezing rain"\newline
\hspace{4em}\hangindent=3em\hangafter=1 O: "Snow squall"\newline
\hspace{4em}\hangindent=3em\hangafter=1 P: "Blizzard"

\textbf{Summary:} Provide a concise summary with a maximum of two sentences that describe the expected severe weather development. Avoid referring to a wide range of severe weather phenomenons and focus on providing the exact type of the severe threat, its intensity, the timing (when in the future or time of day it will occur), and the area likely affected. 

\end{tcolorbox}

\subsection{Mesoscale Discussion Generation Prompt for Case Study}\label{generation prompt}

\begin{tcolorbox}[colback=white, colframe=red, title=Benchmark Prompt Demonstration based on function block \ref{function block generation}, breakable]
First, configure the system prompt based on \textbf{<System prompt>\ref{Function block}} and start with \textbf{<Generation Prompt Instructions>}.\newline
Second, append the function block \textbf{<Generation Question Template>}\newline
Next, select \textbf{<Generation Few-shot>} if needed. \newline
Finally, append the \textbf{<Generation Problem>}. 
\end{tcolorbox}

\begin{tcolorbox}[colback=red!10!white, colframe=red!50!black, title=<Generation Few-shot>]
\textbf{Repeat N times for few-shot:}

\hspace{1em}\texttt{Example ith:}
\begin{enumerate}
    \item Modify \textbf{<Encode Weather Parameters> in \ref{Function block}} given \textbf{<Time of ingredients-based parameters>} and \textbf{<Encoded parameters>}
    \item Instead of providing correct options, the ith example of MCD is provided:
    \begin{itemize}
        \item <Areas Affected> 'Areas Affected' from ith example </Areas Affected>
        \item <Concerning> 'Concerning' from ith example </Concerning>
        \item <Summary> 'Summary' from ith example </Summary>
    \end{itemize}
    Append after \textbf{<Encode Weather Parameters>}
\end{enumerate}

\end{tcolorbox}

\begin{tcolorbox}[colback=gray!10!white, colframe=gray!50!black, title=<Genration Problem>]
Modify \textbf{<Encode Weather Parameters>} given \textbf{<Time of ingredients-based parameters>} and \textbf{<Encoded parameters>}

\hspace{1em}\texttt{Output format:}
\begin{enumerate}
    \item <Area affected>Area\_Affected:</Area affected>
    \item <Concerning>Concerning:</Concerning>
    \item <Summary>Summary:</Summary>
\end{enumerate}
\end{tcolorbox}

%% file: appendix_case_study.tex
\section{Case Study}

\subsection{Grading Rubrics}\label{rubrics}
\begin{tcolorbox}[colback=white!95!gray, colframe=black, title=Grading Rubrics, breakable]

\subsubsection*{1. Areas Affected}
\begin{itemize}
    \item \textbf{Score 0: No Match} \\
    The VLM-generated areas have no overlap with the ground truth areas, indicating a complete mismatch.
    \item \textbf{Score 1: Partial Match} \\
    The VLM-generated areas partially match the ground truth areas, showing some geographical overlap but significant discrepancies in the specific regions mentioned.
    \item \textbf{Score 2: Close Match} \\
    The VLM-generated areas closely match the ground truth areas, with minor differences that do not significantly alter the geographical context of the discussion.
    \item \textbf{Score 3: Exact Match} \\
    The VLM-generated areas exactly match the ground truth areas, showing perfect alignment in the geographical regions discussed.
\end{itemize}

\subsubsection*{2. Concerning (Severe Potential and Probability of Watch Issuance)}
\begin{itemize}
    \item \textbf{Score 0: Incorrect Concerning and Probability} \\
    The VLM-generated concerning and probability of watch issuance are both incorrect and do not align with the ground truth or the established relationship chart.
    \item \textbf{Score 1: Partially Correct Concerning or Probability} \\
    Either the concerning or the probability of watch issuance aligns with the ground truth, but not both. The other element is incorrect or misaligned with the severity chart.
    \item \textbf{Score 2: Correct Concerning and Probability, Minor Contextual Error} \\
    Both the concerning and probability of watch issuance are correct and align with the ground truth. There may be minor errors in how these elements are contextually integrated with the rest of the discussion.
    \item \textbf{Score 3: Fully Correct and Contextually Insightful} \\
    Both the concerning and probability of watch issuance are correct, perfectly align with the ground truth, and are well integrated within the context of the entire mesoscale discussion.
\end{itemize}

\subsubsection*{3. Summary Consistency with Weather Analysis Images}
\begin{itemize}
    \item \textbf{Score 0: No Consistency} \\
    The summary provided by the VLM has no consistency with the weather analysis images, showing a complete lack of understanding or correlation.
    \item \textbf{Score 1: Low Consistency} \\
    The summary shows low consistency with the weather analysis images. There are significant discrepancies in the description of weather conditions or expected developments.
    \item \textbf{Score 2: Moderate Consistency} \\
    The summary is moderately consistent with the weather analysis images. Most of the described weather conditions or developments are in line with what is shown in the images, with some minor errors or omissions.
    \item \textbf{Score 3: High Consistency} \\
    The summary is highly consistent with the weather analysis images. It accurately reflects the weather conditions and expected developments depicted in the images, demonstrating a deep understanding of the meteorological context.
\end{itemize}

\subsubsection*{Total Score}
\begin{itemize}
    \item \textbf{0-3 Points: Poor performance}, contain misleading information and significant improvements needed.
    \item \textbf{4-6 Points: Fair performance}, some aspects are well-handled while others need refinement or revision.
    \item \textbf{7-8 Points: Good performance}, minor errors that do not significantly impact the forecasting quality.
    \item \textbf{9 Points: Excellent performance}, accurately reflects the ground truth and shows a good understanding of the meteorological data.
\end{itemize}

\end{tcolorbox}

\subsection{Feedback Guidelines}\label{Guidelines}

\begin{tcolorbox}[colback=white!95!gray, colframe=black, title=Feedback Guidelines, breakable]
\textbf{Confidence Assessment}

\begin{itemize}
    \item How confident do you feel about the VLM's generated mesoscale discussion in terms of accuracy and reliability when compared to the ground truth?
    \item Does the VLM-generated content inspire the same level of trust as the ground truth discussion for decision-making purposes?
\end{itemize}

\textbf{Specificity vs. Broadness}

\begin{itemize}
    \item Would you prefer the VLM to provide more specific types of severe weather predictions, or is a broader range of potential severe weather phenomena more useful?
    \item How does the specificity or broadness of the VLM's severe weather descriptions compare to the ground truth?
\end{itemize}

\textbf{Interpretation of Weather Analysis Maps}

\begin{itemize}
    \item How well does the VLM interpret and integrate weather analysis maps into its mesoscale discussions?
    \item Are there any indications that the VLM misunderstands or misrepresents the data from the weather analysis maps?
\end{itemize}

\textbf{Hallucinations and Factual Accuracy}

\begin{itemize}
    \item Did you notice any instances where the VLM 'hallucinated' details, meaning it generated information not supported by the ground truth or weather analysis maps?
    \item How does the factual accuracy of the VLM's discussion compare to the ground truth?
\end{itemize}

\textbf{Consistency and Contextual Understanding}

\begin{itemize}
    \item How consistent is the VLM's mesoscale discussion in terms of the overall context provided by the ground truth and weather analysis images?
    \item Does the VLM demonstrate a contextual understanding of the severe weather situation, or does it seem to generate isolated facts without a coherent narrative?
\end{itemize}

\textbf{Language and Clarity}

\begin{itemize}
    \item Is the language used by the VLM clear and professional, akin to what would be expected in a ground truth mesoscale discussion?
    \item Are there any issues with the readability or technical terminology used by the VLM?
\end{itemize}

\textbf{Appropriateness of Recommendations}

\begin{itemize}
    \item How appropriate are the VLM's recommendations or predictions regarding watch issuance compared to the ground truth?
    \item Does the VLM align its recommendations with the established probability chart for watch issuance?
\end{itemize}

\textbf{Temporal Aspects}

\begin{itemize}
    \item Does the VLM accurately reflect the timing aspects of the severe weather events (e.g., "in the next 1-2 hours") as compared to the ground truth?
    \item How well does the VLM handle the dynamic nature of weather forecasting in its discussions?
\end{itemize}

\textbf{Improvement and Learning}

\begin{itemize}
    \item Based on the comparison, what improvements would you suggest for the VLM to better align with the ground truth mesoscale discussions?
    \item Are there any patterns in errors or inaccuracies that could indicate specific areas where the VLM needs further training or refinement?
\end{itemize}

\textbf{Overall Evaluation and Preference}

\begin{itemize}
    \item Overall, how does the VLM-generated mesoscale discussion measure up to the ground truth?
    \item Given the choice, do you think VLM-generated mesoscale discussion could aid forecasters in the process of writing formal mesoscale discussion?
\end{itemize}
\end{tcolorbox}

\subsection{Expert review procedure}\label{expert-review-procedure}
Each expert is given a grading rubric and feedback guidelines before starting the evaluation. They also receive a grading spreadsheet to record the scores for each model's output. The experts are provided with MCDs generated by both the 3-shot GPT-4O and the fine-tuned Llama2, based on the weather conditions and issuance time of the selected MCD. Additionally, the selected MCD and weather conditions are provided to help the experts assess the performance of the VLMs. Experts are allowed to use additional information for their evaluation, which can include earlier MCDs or other weather conditions not available as input. The models are anonymized, with GPT-4O referred to as 'model 1' and fine-tuned Llama2 as 'model 2'. Experts are also encouraged to provide detailed feedback along with their grades.

\subsection{Rating distribution}\label{appdx: rating distribution}
\begin{figure}[ht]
\begin{center}
\centerline{\includegraphics[width=1\columnwidth]{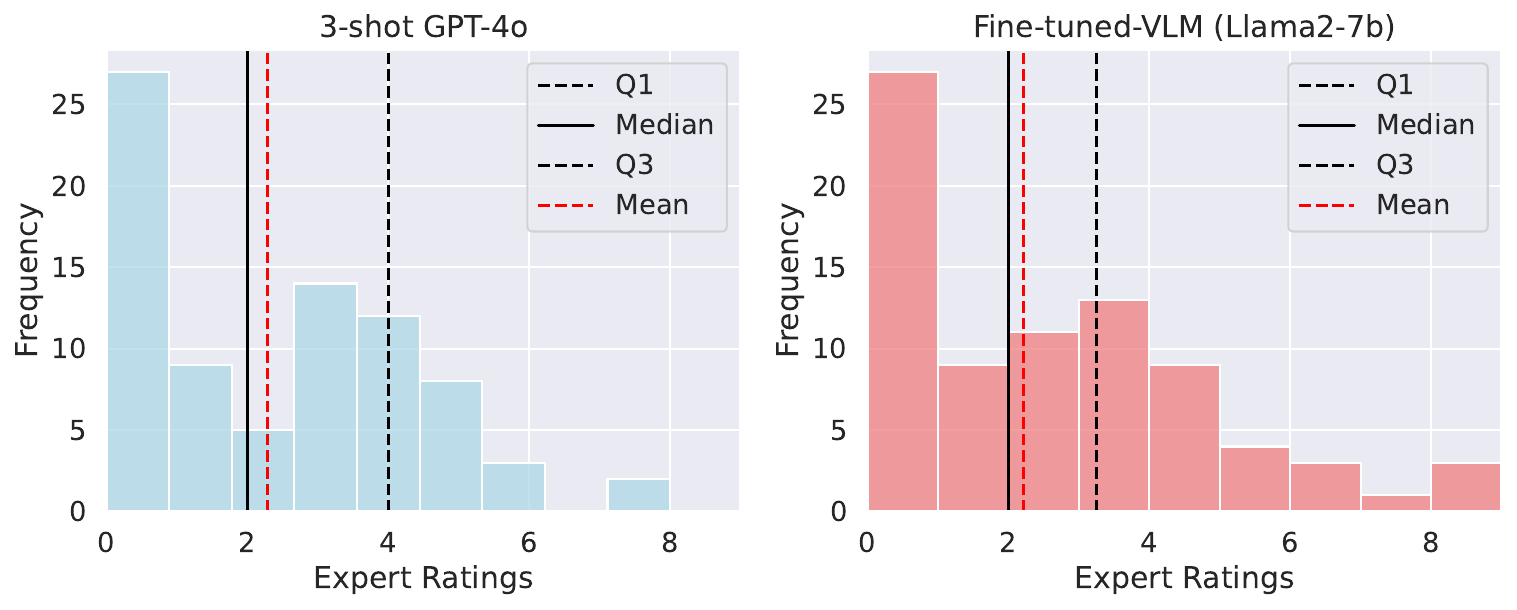}}
\caption{Distribution of Expert Ratings for 3-shot GPT-4o and Fine-tuned-VLM (Llama2)}
\label{fig: expert ratings}
\end{center}
\end{figure}
The rating distribution is shown in Figure \ref{fig: expert ratings}. A total of 80 samples of ratings were collected from 4 experts. The overall performance of Fine-tuned-VLM (Llama2-7b) is very similar to 3-shot GPT-4o. Both models have a similar distribution from scores of 0 to 2, with 3-shot GPT-4o having more examples rated higher in the range of 3 to 5. However, Fine-tuned-VLM (Llama2-7b) is able to get slightly more high ratings from 6 to 9.

\subsection{Examples}\label{cases-appendix}
\begin{figure}[ht]
\begin{center}
\centerline{\includegraphics[width=1\columnwidth]{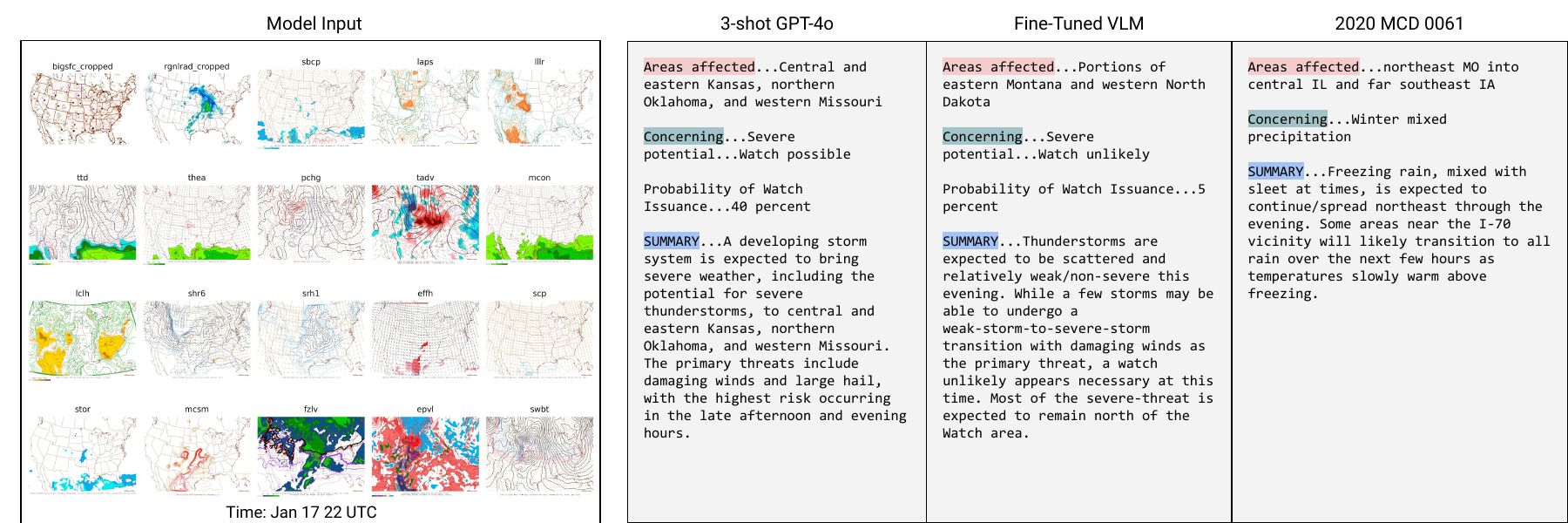}}
\caption{Similar to Figure \ref{fig:case study demo} but for case 2020 MCD 0061}
\label{2020MCD0061}
\end{center}
\vskip -0.2in
\end{figure}

\begin{figure}[ht]
\begin{center}
\centerline{\includegraphics[width=1\columnwidth]{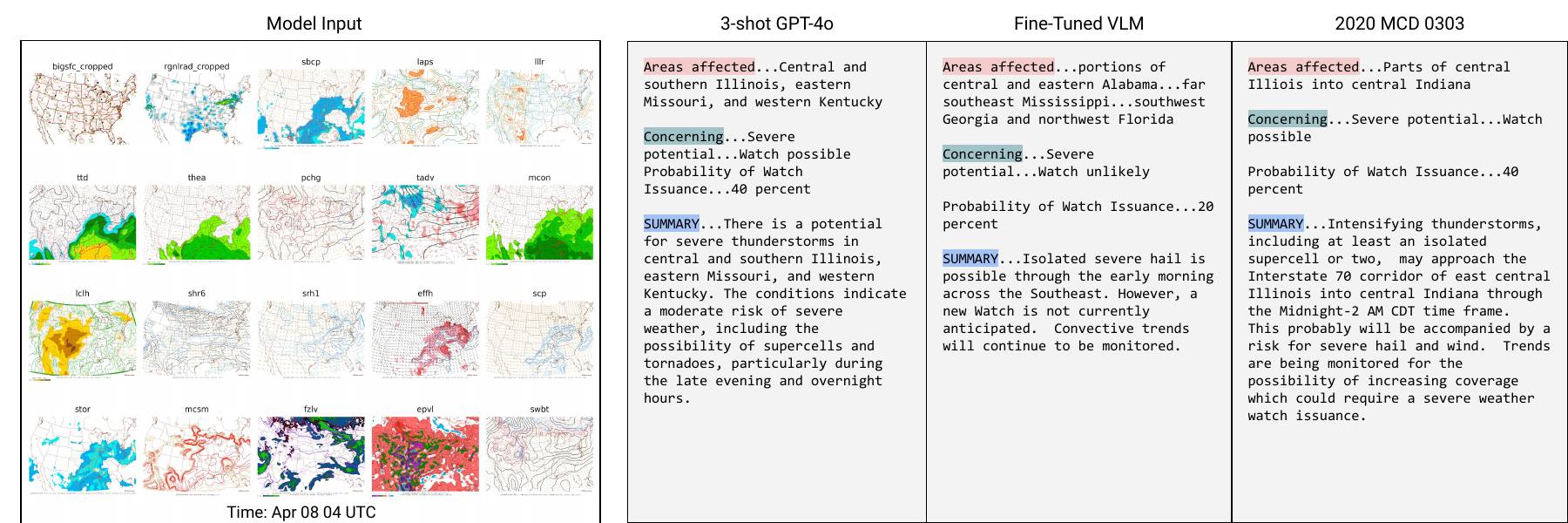}}
\caption{Similar to Figure \ref{fig:case study demo} but for case 2020 MCD 0303}
\label{2020MCD0303}
\end{center}
\vskip -0.2in
\end{figure}

\begin{figure}[ht]
\begin{center}
\centerline{\includegraphics[width=1\columnwidth]{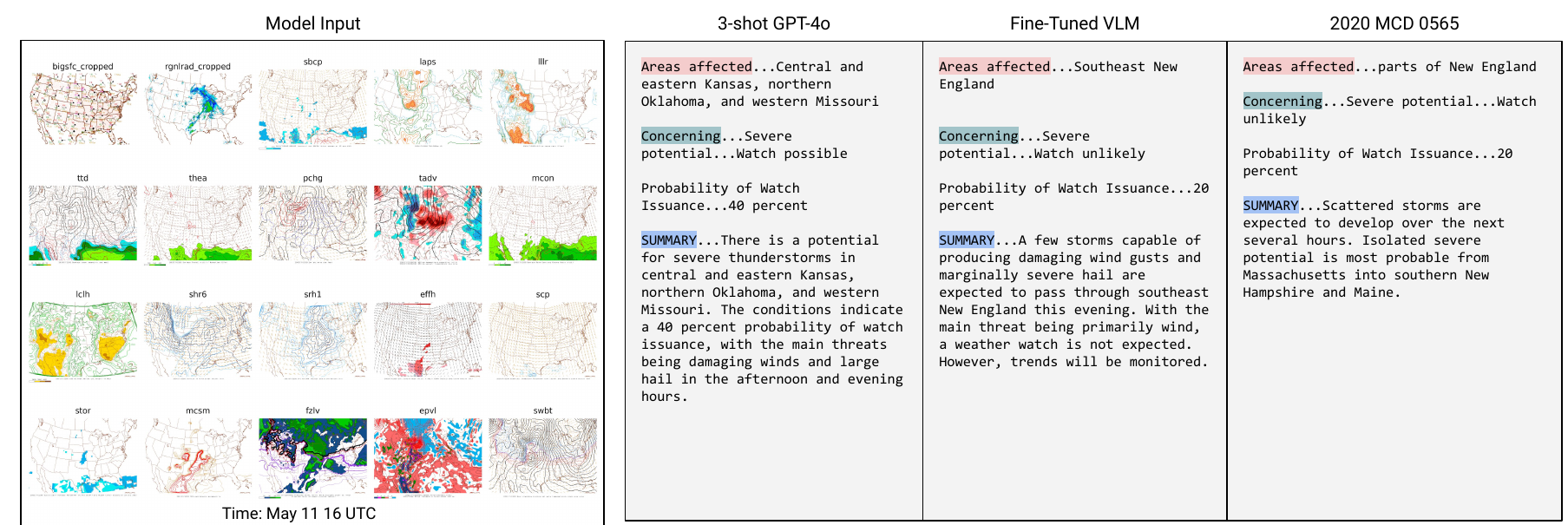}}
\caption{Similar to Figure \ref{fig:case study demo} but for case 2020 MCD 0565}
\label{2020MCD0565}
\end{center}
\end{figure}

\begin{figure}[ht]
\begin{center}
\centerline{\includegraphics[width=1\columnwidth]{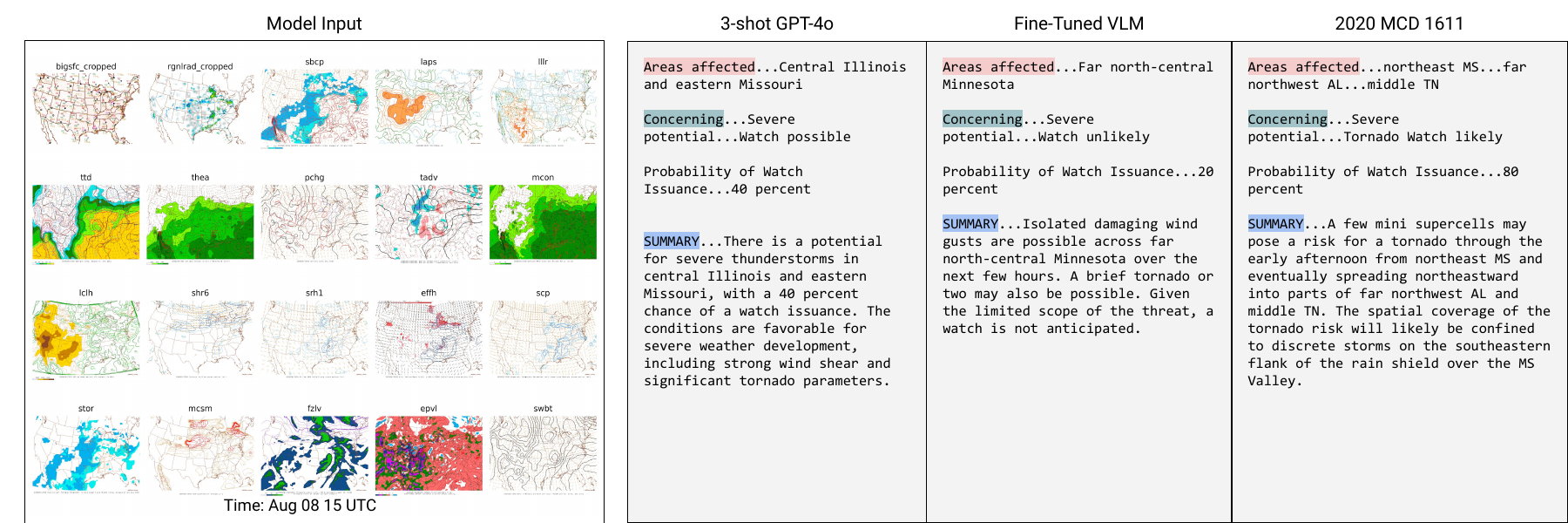}}
\caption{Similar to Figure \ref{fig:case study demo} but for case 2020 MCD 1611}
\label{2020MCD1611}
\end{center}
\end{figure}

\begin{figure}[ht]
\begin{center}
\centerline{\includegraphics[width=1\columnwidth]{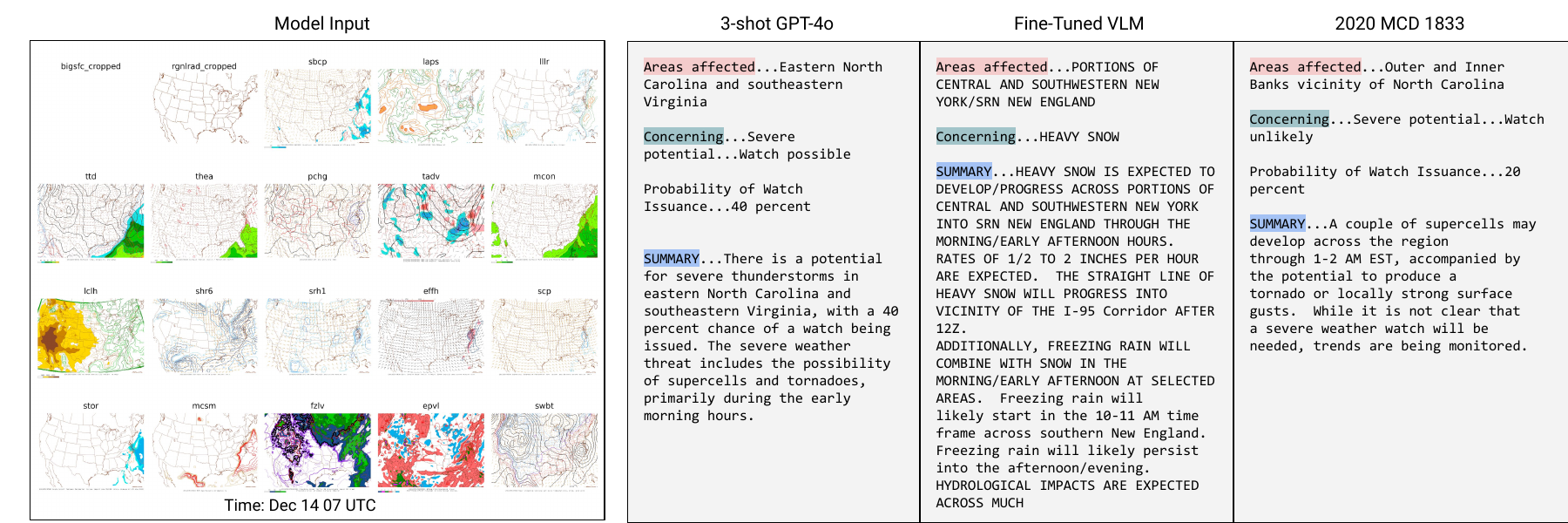}}
\caption{Similar to Figure \ref{fig:case study demo} but for case 2020 MCD 1833}
\label{2020MCD1833}
\end{center}
\end{figure}

%% file: supplementary.tex
\section*{Dataset Details, Release, and Licensing for WeatherQA}

We provide our WeatherQA dataset, designed to explore the potential of multimodal model for understanding and reasoning about severe weather phenomena specifically thunderstorms and hazardous winter conditions across contiguous United States in the upcoming hours. The dataset is accessible through \url{https://github.com/chengqianma/WeatherQA}

\subsection*{Release and Licensing}

Our dataset is released under the CC0 license, ensuring open access and broad usability. Here is how we publicly release our dataset:

\subsubsection*{Code and Data Availability}

We include all the code and data necessary for utilizing the dataset. This includes:

\begin{itemize}
    \item \textbf{WeatherQA Dataset}: Includes unique identifiers for each Mesoscale Discussion sample, an array of file paths to 20 different weather parameter images, detailed annotations describing weather conditions and threats, and the date and time of the weather parameter images in UTC.
    \item \textbf{Mesoscale Analysis (image) Dataset}: Includes 20 images per sample, including 18 weather parameter images, one surface observation image, and one composite radar reflectivity image, organized by year and weather parameters in the './md\_image' directory, with each image file named following a specific convention.
    \item \textbf{Test Data}: Includes system prompts, prompt templates, weather parameter descriptions, and test data samples with unique identifiers, weather parameter paths, timestamps, choices, and correct answers for areas affected and concerning classifications.
    \item \textbf{Code}: Includes test scripts designed to run benchmarks on various multimodal models (Claude, Gemini, and GPT-4) that are used in this study. The code for fine-tuning the Vision-Language Model Llama, as well as scripts for performing inference tasks, are also included. Additionally, the pipeline for data curation is also provided. 
    \item \textbf{Expert Comments}: Includes the case study feedback and the grades from the experts.
\end{itemize}